# Live American Sign Language Letter Classification with Convolutional Neural Networks

Kyle Boone, Ben Wurster, Seth Thao, and Yu Hen Hu
University of Wisconsin - Madison

Abstract:

This project is centered around building a neural network that is able to recognize ASL letters in images, particularly within the scope of a live video feed. Initial testing results came up short of expectations when both the convolutional network and VGG16 transfer learning approaches failed to generalize in settings of different backgrounds. The use of a pre-trained hand joint detection model was then adopted with the produced joint locations being fed into a fully-connected neural network. The results of this approach exceeded those of prior methods and generalized well to a live video feed application.

## Introduction:

Hundreds of thousands of people across this country rely on American Sign Language (ASL) everyday for communication. This project attempts to generate live video feed classification of ASL letters. Potential applications for this include generating live subtitles on videos of sign language users, providing a space for ASL learners to practice with feedback, or using it in a Google Translate like app that translates live video feed from a user's camera.

It is not surprising that there are existing software applications to effectively translate from ASL to English. An example of this is SignAll [1], which reports to have the largest database of ASL in the world. They use a live video feed that combines hand signals, facial expressions, and body language to fully translate ASL into English. Things like facial expressions and body language can help distinguish between words of different connotations based on the emotions the person is showing. This model is obviously much more advanced than anything producible over the course of a semester, but it does show the potential of neural networks in this field.

SignAll is far from the only group working on this project. Results from multiple different state of the art models can be found reported by Madhiarasan [2]. A wide variety of approaches have been used with a wide range of success. The most successful models incorporate many factors such as facial expression recognition (as mentioned above), but unsurprisingly there is a tradeoff between accuracy and speed. The most accurate models are reported to have accuracies of over 98%. Even though this project would only focus on letter classification, it should be noted that this is one of the most useful areas of recognition in sign language, as if a word doesn't have an ASL representation, it is often a simple substitute to spell it out letter-by-letter. There are over an order of magnitude more English words than ASL signs, so this scenario comes up quite often.

While these state-of-the-art models are fascinating, they tackle a much more complex problem in attempting to classify all words rather than just letters. Furthermore, the size of the datasets used by these models is hardly comparable to the size of a dataset practical for use in a project for this course. Because of this, a more suitable baseline result completed by Bheda and Radpour was found [3]. This project focuses on the classification of stationary images of ASL letters. They reported an accuracy of 82.5% on a premade dataset and a 66.7% accuracy on their own custom made dataset. Their model worked by using pre-made software to isolate the hand by blacking out the background, and then applying a convolutional neural network to the resulting image. Examples of some of their images are shown in Figure 1.

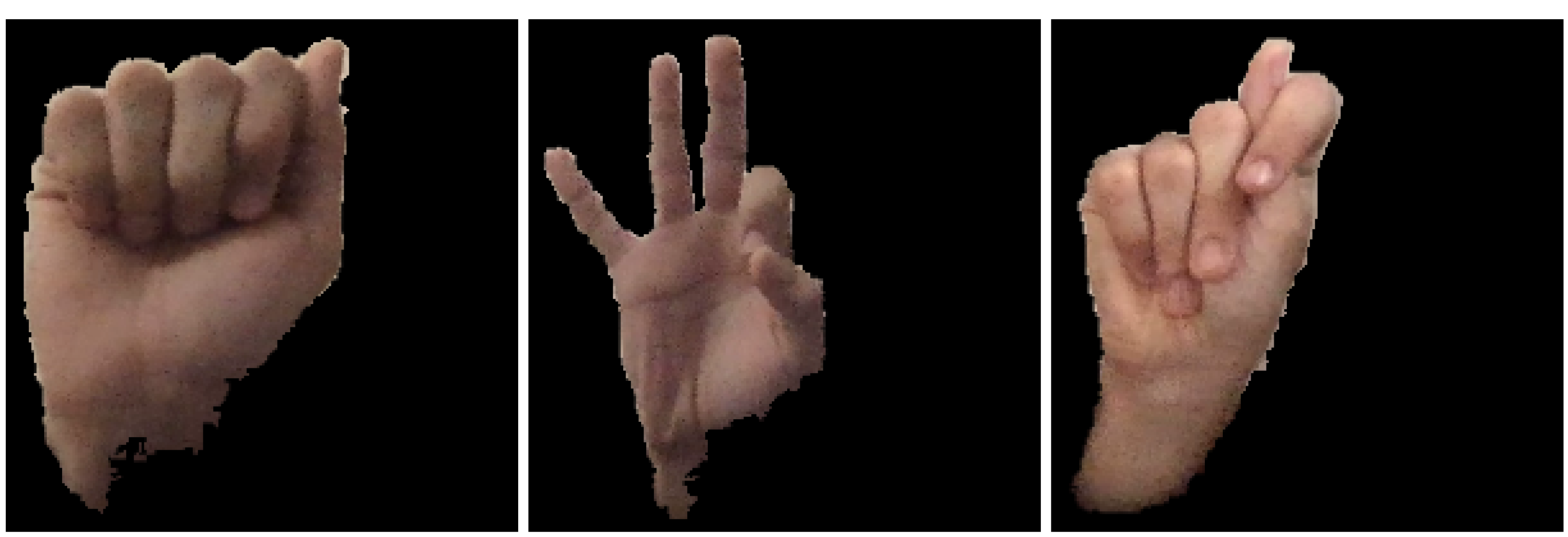

**Figure 1.** Signed A, F, and T (left to right) from the Bheda and Radpour dataset.

It should be noted that their accuracy claim was tested and found to be inaccurate. When tested on their custom made dataset, an accuracy of only 41.7% was found, and when tested on a custom made dataset created for our project (discussed later), an accuracy of only 12.5% was found. It should be noted that when tested on our custom dataset, there were some mismatches in letter symbols, such as a different sign for “T” (discussed later) that contributed to the lower accuracy. Their GitHub is located at [4]. Despite the fact that their model tested much worse than reported accuracy wise, a goal of 85% accuracy was set to beat their reported result.

## Data:

Multiple different datasets were used throughout the creation of the model. The first dataset was obtained from Kaggle [5]. However, there are some issues with this dataset, largely with the fact that some letters (most notably T) are not accurate to ASL. While this is problematic, the main goal of this project is to demonstrate the potential of a neural network for letter classification, which can still be demonstrated regardless of the accuracy of the training dataset as long as the network can distinguish a variety of hand signs. Because of this, the benefits of this dataset were deemed to outweigh the drawback of a few inaccurate letters.

This dataset was chosen largely because of its sheer size at 87,000 total images. These images were split evenly among 29 different classes, for a total of 3,000 images per class. The classes were the 26 English letters as well as a delete character, a space character, and a nothing character (no hand was present). These images were all 200x200 pixels and were all color images.

The dataset technically had a designated testing dataset, but this only contained 28 images, one of each class except for delete. Due to the small size of this explicit testing dataset, it was discarded, and instead, the 87,000 image dataset that was originally labeled as “training” was partitioned into 80% training, 10% validation, and 10% testing subsets. These splits were not stratified since with a dataset containing this many images per class, even without explicit stratification done, the testing dataset should statistically wind up with a satisfactory number of images from each class.

For preprocessing the data, these images were converted to grayscale. The reasoning for this was that all of the images had very similar backgrounds, so in theory the neural network could simply learn to ignore this specific background to isolate the hand. While that would work well for the Kaggle dataset, it would not generalize to other backgrounds.

Beyond this, there was one further step of data augmentation that was done: brightening the images. This was a necessary method to use on the grayscale dataset, since many images were quite dark as a result of the conversion. Some examples of these images from the augmented dataset are shown in Figure 2.

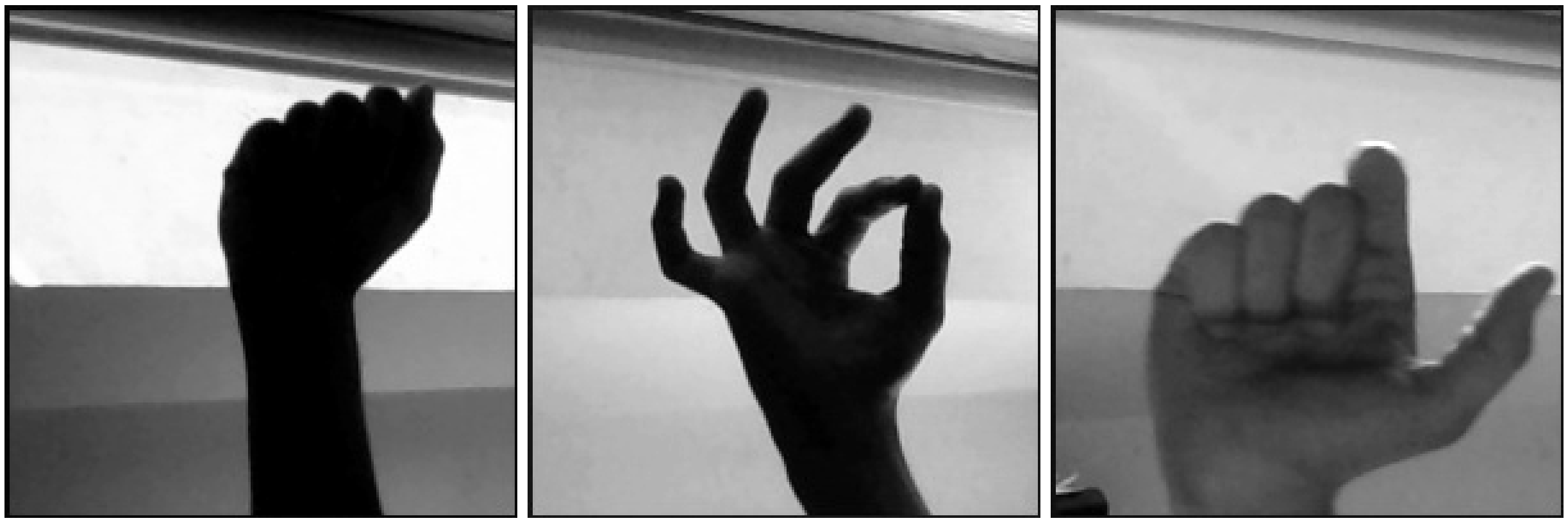

**Figure 2.** Signed A, F, and T (left to right) in the Kaggle dataset.

After training on this dataset, the model was found to be unable to generalize to different backgrounds, so a custom dataset with multiple different backgrounds was made. To stay consistent with the Kaggle dataset, the same signs were used as they had for each letter and special character. In theory, this was so that both datasets could be used for training and could be compared. This included the previously mentioned characters that were not accurate to ASL such as their letter “T”.

A total of 87,000 custom images were taken. Once again, this was distributed evenly among the 29 classes for a total of 3,000 images per class. The images were 200x200 pixels and were in color.

Like with the Kaggle dataset, data was again partitioned with 80% going into a training set, 10% to validation, and 10% to testing. Again, due to the large number of samples per class, the data partitioning was not stratified as it was deemed unnecessary since there would be a relatively even distribution regardless. Unlike the Kaggle dataset, these images were not converted to grayscale since the backgrounds were more varied. Due to the variation, it was deemed to be less likely for the neural network to simply pick up on one specific background due to its colors.

More data augmentation was done on these images than the Kaggle dataset images. Random horizontal flips were implemented so that the network would work for both left and right handed people. Furthermore, random brightness and contrast changes were performed in an effort to help the network generalize to different lighting and camera conditions. Things like random rotations were not done since those could change the label that certain letters in ASL should be assigned with. This is most obvious with the letters “I” and “J”, which differ in the dataset by only rotation. It was determined that natural variances due to human error would be enough to capture a reasonable range of rotations without the risk of having overlap between letters. An example of some of the letters from this new dataset are shown in Figure 3.

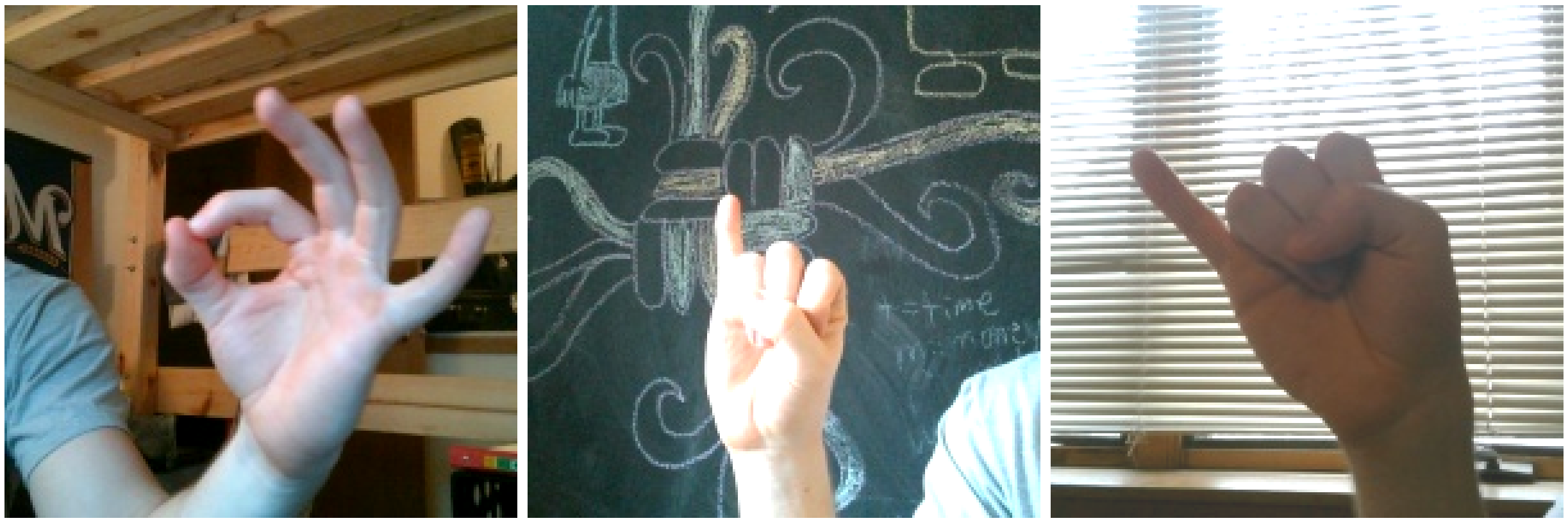

**Figure 3.** Signed F (flipped horizontally), I, and J (left to right) in the custom dataset.

## Tasks Performed:

In this project, three separate model architectures with varying degrees of complexity were used. The first model built was a basic convolutional neural network. The actual architecture of the model is shown on the right. As an overview, the model consisted of three convolutional layers followed by three fully connected layers. Each of the convolutional layers except for the last one was followed by a max pooling layer. The last convolutional layer drastically reduces the dimensionality of the feature vectors in an effort to decrease the number of trainable parameters. In total, this model had 18,967 trainable parameters. With a dataset size of 87,000 images, this was deemed as a number of parameters that was not too high in terms of overfitting risk. Adam was used for the optimizer due to its high convergence rates on every problem seen during this course, and categorical cross entropy was used for loss due to this being a classification problem. For this initial model, the Kaggle dataset, with the augmentations described in the above section, was used.

A grand total of 120 epochs were used in training this model. The batch size for these epochs started at 50 for the first 30 epochs, went to 300 for the next 30 epochs, and finally ended at 600 for the last 60 epochs. It was found that the smaller batch epochs improved accuracy quicker at the beginning of training; however, these smaller batches were unable to achieve as high of accuracy in the long term as larger batch sizes, which is why this approach was used.

The second approach used was transfer learning. In an effort to prevent overfitting, a base model with a relatively low number of output dimensions compared to other transfer learning models was used, that model being VGG16. The custom dataset with the augmentations described in the data section was used for this model to try to help it generalize to new backgrounds. The VGG16 model expects an input of 224x224 color images, so the images were reshaped to 224x224 before any operations were done to them (including the data augmentation). After the data augmentation, the images were preprocessed using the built in VGG16 preprocessing method. After this, the images were fed through the VGG16 model, giving outputs of the shape 7x7x512. Global average pooling was applied to get this down to 1x1x512, which was then flattened to simply get a 512 dimensional feature vector. A single fully connected layer was attached to the end of this to get the predicted outputs. Overall, due to the relatively low number of output channels from VGG16, there were a total 14,877 trainable parameters since all of the VGG16 layers were frozen. This number was deemed acceptable in its risk to overfit. Once again, Adam was used for the optimizer while categorical cross entropy was used for loss for the same reasons as stated before. The architecture of this model is shown on the right. The top layer represents the entirety of the VGG16 base network. Due to the high validation and testing accuracy that was achieved in the initial model, the same number of epochs was used with the same minibatch size distribution.

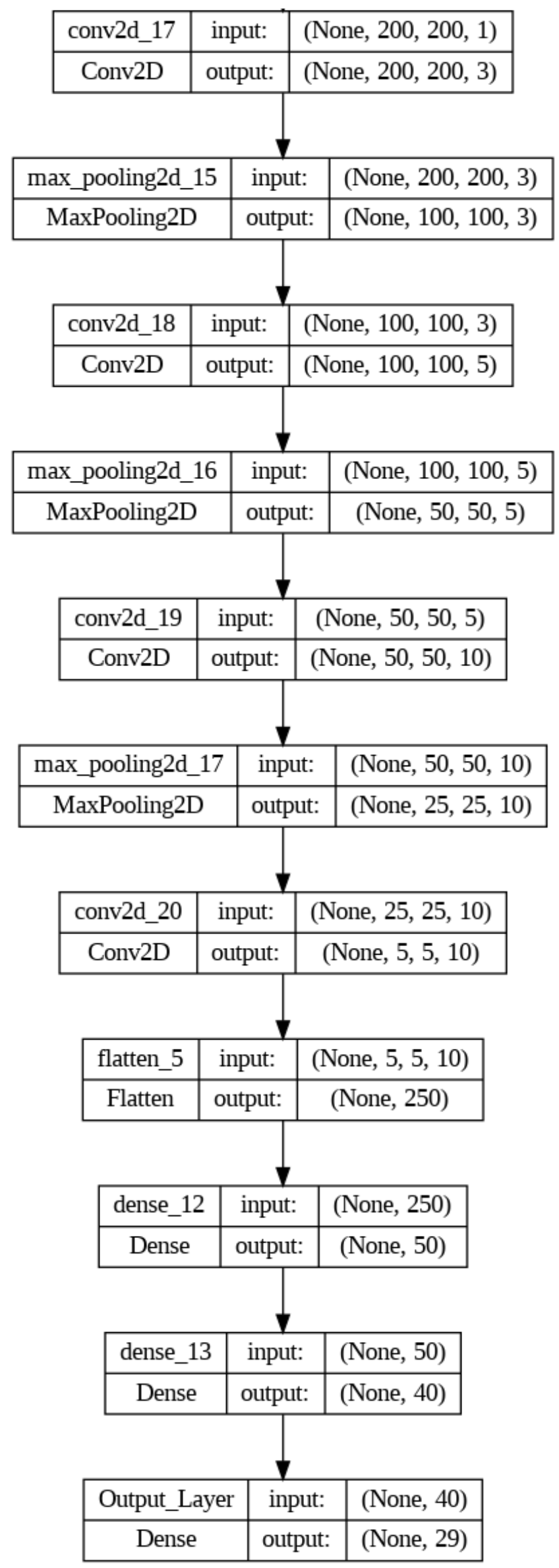

**Figure 4:** Initial Model

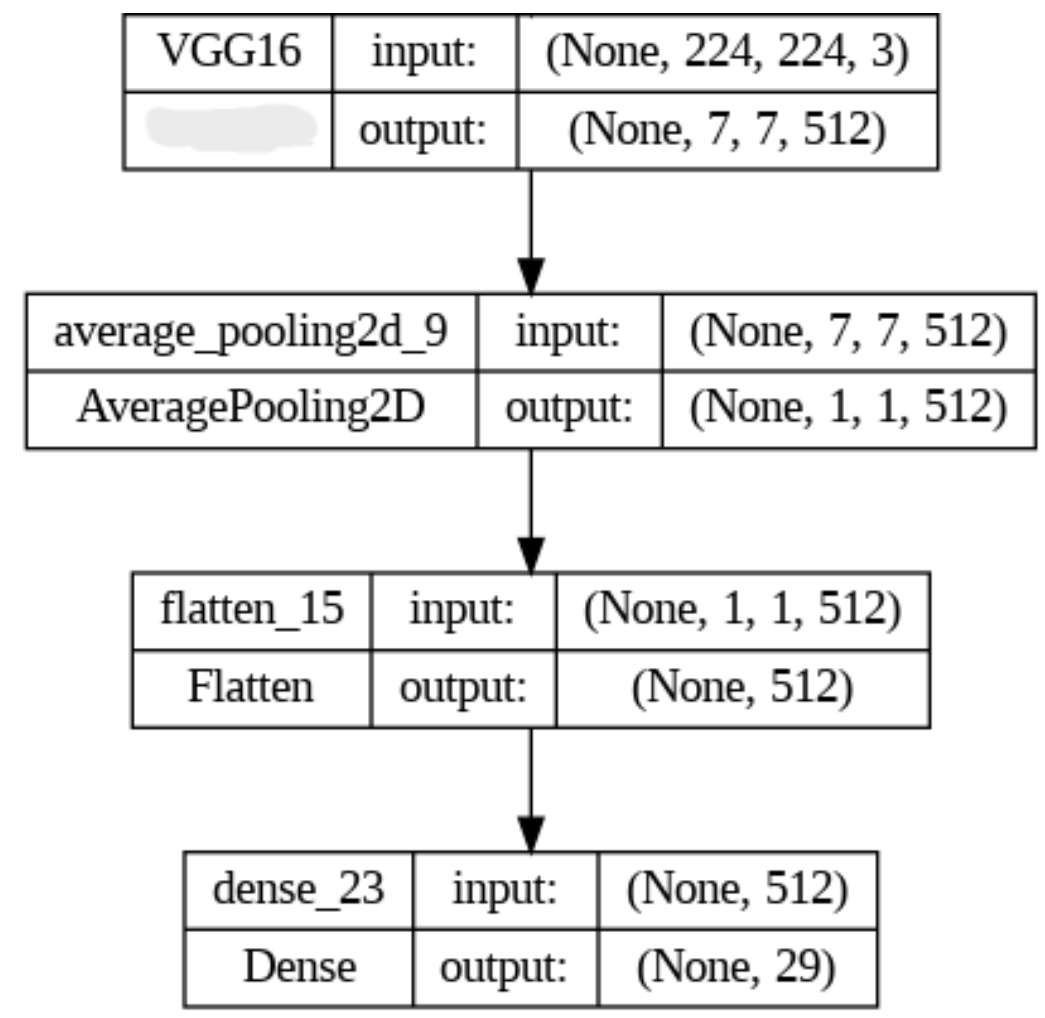

**Figure 5:** Transfer Learning Architecture

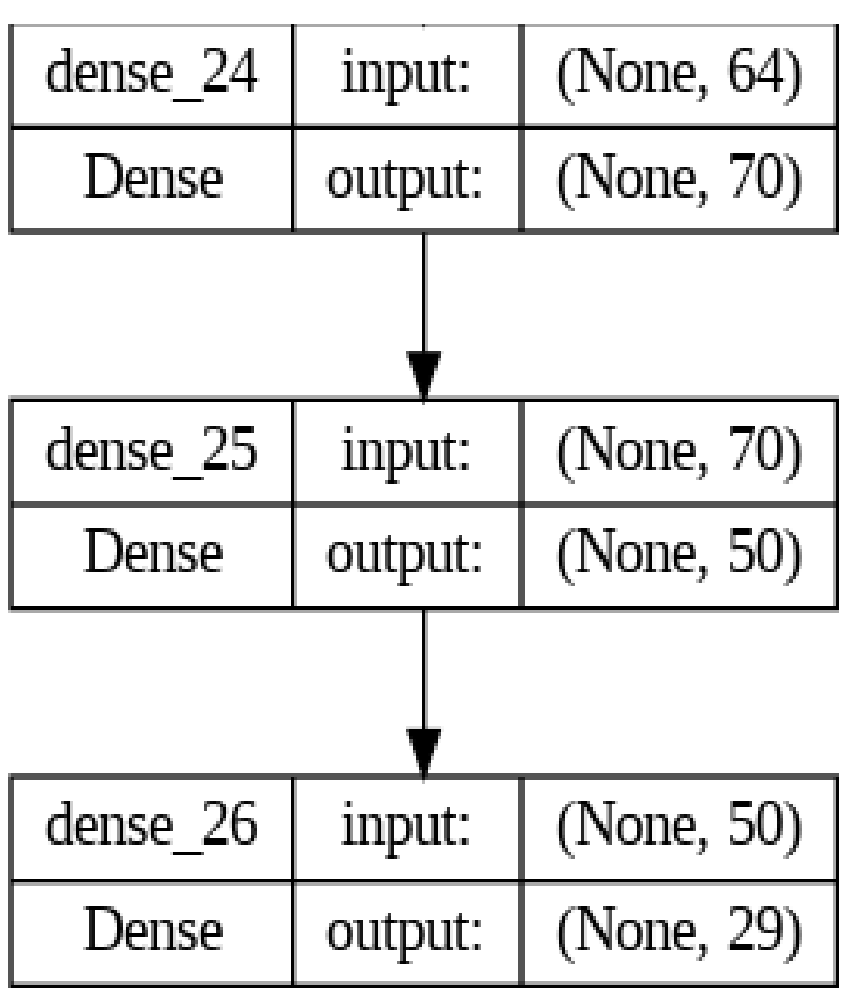

**Figure 6:** MediaPipe Architecture

The final model that was built used, as per the recommendation of Professor Hu, a prebuilt hand joint detector to give the three dimensional locations of 21 unique hand features before attempting to map these joint locations to letters with a fully connected, non convolutional neural network. MediaPipe [6] is an open-source project developed by Google in 2020 to perform complex image recognition tasks including background segmentation, pose detection, face tracking, and hand detection. The last of these was incredibly pertinent to the problem at hand. The built-in hand detector works on any image size, so no reshaping was required and the images remained 200x200. The custom data set was used, and data augmentations were applied to the images (as described in the data section) before feeding the images into the hand detector. The most important augmentation technique for this method was the horizontal flip, so that the neural network learns to recognize both left and right handed signs. The other data augmentation methods act more as a stress test for MediaPipe to see when it stops being able to detect hands, since a brightness or contrast change should not actually impact joint locations if the hand is correctly identified.

If MediaPipe manages to detect a hand, 21 joint locations are given in three dimensional coordinates, for a total of 63 coordinates. Additionally, a binary value is given to signify whether the hand detected was the left or right hand. Combining these, the output from MediaPipe that was used as the input to the neural network was a 64 dimensional feature vector. With these inputs, a three layer, fully connected neural network was built to attempt to classify the signs. The architecture is shown in Figure 6. A total of 9,579 parameters were trainable, which was deemed an acceptable number. Once again, Adam was used for the optimizer and categorical cross entropy was used for loss. The same epoch number was used with the same distribution of batch sizes.

As mentioned in the introduction section, the baseline model to compare against uses pre-built software to isolate the hand, and then pass this isolated hand with a blacked out background as an image into a convolutional neural network for classification. None of the models built here isolate the hand as an image, so all of these models work in a fundamentally different way from their project. The closest model to theirs is the MediaPipe model, but this model does not isolate the hand by blacking out the background, but instead gets joint locations, changing the remaining problem away from an image classification problem.

For each model, the epoch with the highest validation accuracy was saved. Result analysis for all of the models was done with the seaborn confusion matrix, examples of which are shown in Figures 7-9. As well as this, simple numerical results such as test and validation accuracy were recorded for all of the saved models.

The MediaPipe-based model resulted in a highly generalizable model, so with such a model successfully trained to perform the classification of hand positions, a live feed video application was set out to be developed. In order to do this, MediaPipe was once again used to obtain the 21 joint positions and generate the 64-dimensional feature vector for each frame. This was then fed into the neural network with the saved weights from training preloaded to perform classification.

In order to increase the accuracy of the values produced, a confidence bound, initially set at 10, was introduced to represent the number of consecutive frames that must be seen and predict the same letter before a letter would be considered an intentional action. This served a few purposes. The first of these is an anomaly filter. The neural net can make incorrect predictions, but over a long time and many guesses, this becomes increasingly improbable. A secondary effect of this was delay between hand movements. As the user switches their hand position to sign another letter, the neural network can predict incorrect values from intermediate hand positions. By taking consecutive frames into account, this issue is alleviated.

Once a prediction is deemed to be confidently determined, the program types out the letter in the terminal and stores the overall string value produced over the course of program execution. This allows the user to type and visualize a string of characters over time as well as go back, should any mistakes occur, with the delete functionality.

The delete functionality is a special operation that occurs when the delete sign is detected. This reverses the output of the string and removes one character at a time from the stored string value. This undo functionality makes for a quite versatile and usable program to print out any string value exclusively with signed letters while limiting and addressing mistakes that could occur in the process of using the program. This program was tested manually by signing out multiple different words or phrases and ensuring that the model could accurately tell what was being said. A confusion matrix was generated to test the results after manually testing each character.

## Results:

Each of the models discussed above was tested extensively to determine which would work best for a live video feed application. The test accuracy, validation accuracy, and confusion matrix for each model was calculated using the partitioned data mentioned in the data section. For the confusion matrix, the testing subset was used each time.

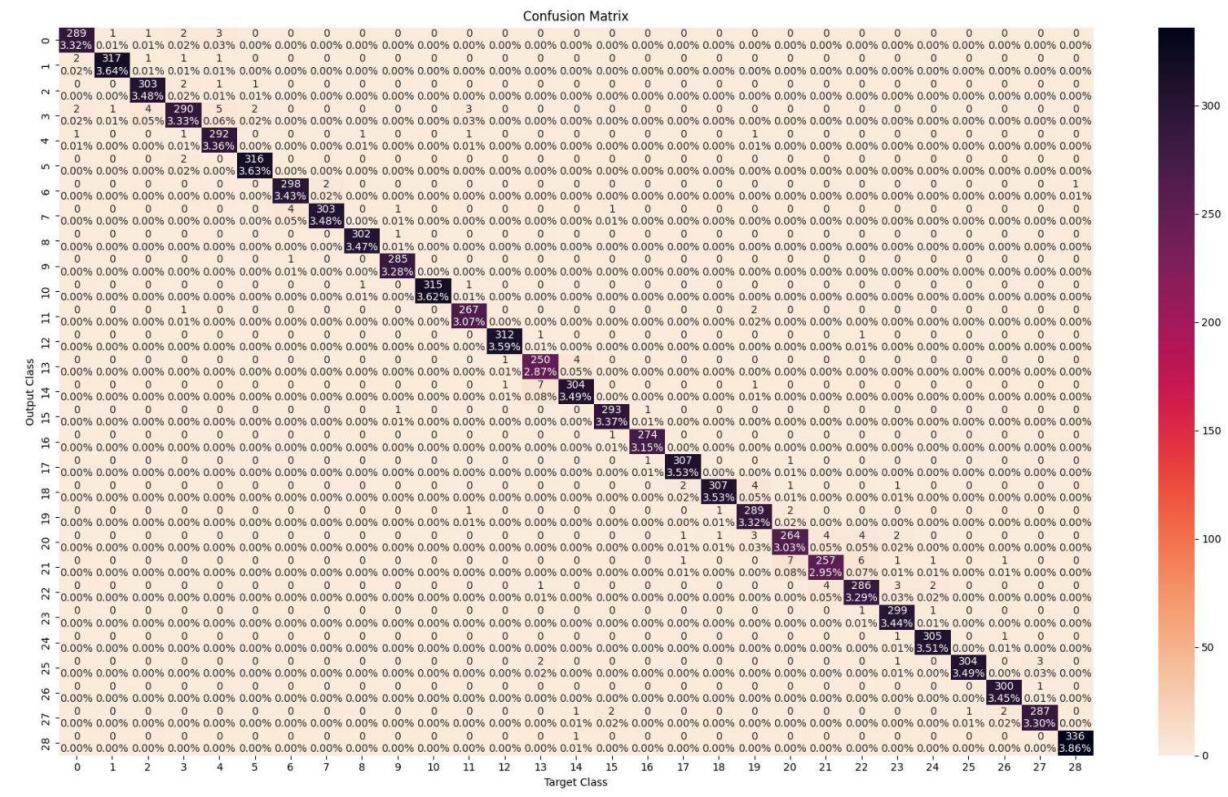

**Figure 7:** Initial Model Confusion Matrix

The initial convolutional neural network was trained with the Kaggle dataset. Overall, it had a validation accuracy of 98.58% on its best epoch and a testing accuracy of 98.29%. The confusion matrix for this model is shown on the right in Figure 7. The confusion matrix for such a model with 29 outputs will obviously be too dense to make out numbers, so refer to the colorbar to get a general idea of performance. The x axis corresponds to the target label, and the y axis corresponds to the output label. While the confusion matrix and reported accuracies sound very positive, the model was unable to generalize to any images it was given with different backgrounds. This was determined after attempts to get predictions on new images with unique backgrounds failed.

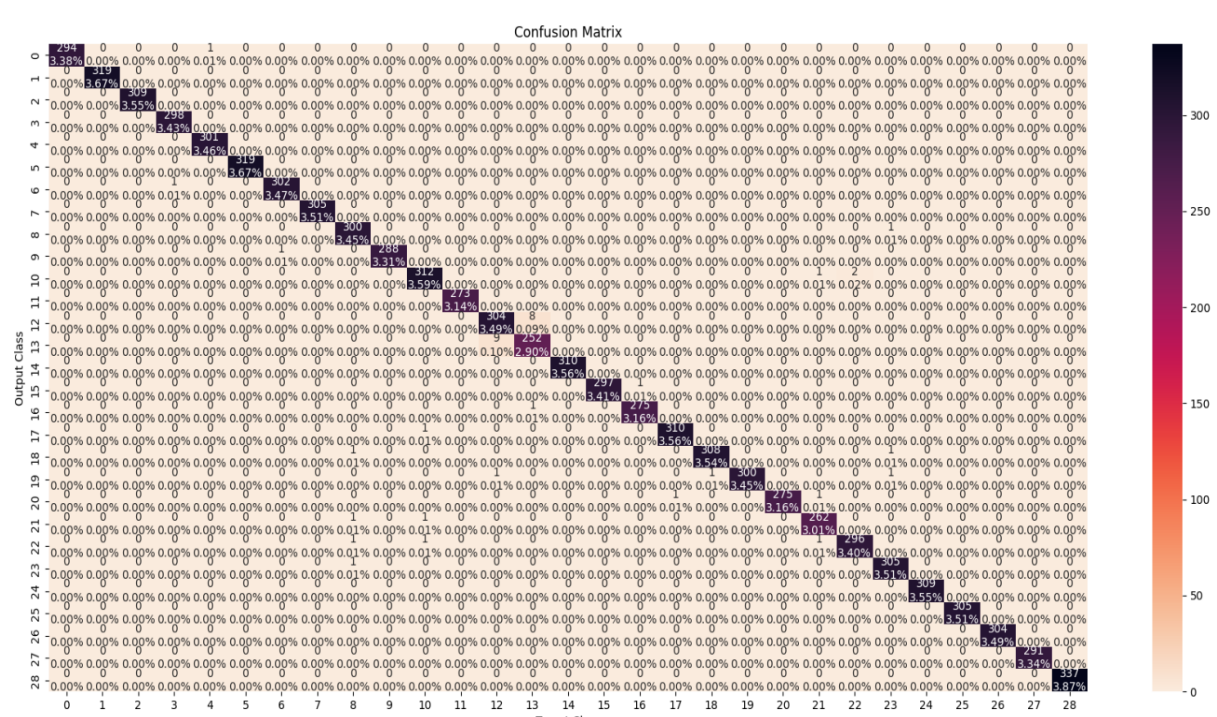

**Figure 8:** Transfer Model Confusion Matrix

The transfer model was trained and tested using the custom made dataset. Overall, it had a validation accuracy of 99.66% on its best epoch and a testing accuracy of 99.54%. The confusion matrix for this model is shown on the left in Figure 8. Once again, the accuracies and confusion matrix make this model look very promising; however, just like the initial model, it was unable to generalize to new backgrounds. Even though this model was trained on images with a variety of backgrounds, it was not able to generalize to images with unique and unseen backgrounds.

Finally, the MediaPipe model was trained and tested using the custom made dataset. Overall it had slightly lower accuracies compared to the transfer model, with a 98.80% validation accuracy on its best epoch and a testing accuracy of 99.00%. The confusion matrix is shown on the right in Figure 9. One thing that can be immediately noticed is that there is more color variation along the diagonal. The reasoning for this is that many samples simply did not have a detected hand. If this was the case, the output from MediaPipe could not be fed into the 64-dimensional input for the neural network. Because of this, samples that did not have a detected hand were not fed to the neural network at all. As can be seen, this did not impact all letters equally. “M” and “N” are among the most impacted by this, and as no surprise, the last column corresponding to the “Nothing” character is essentially completely blank. Overall, for characters besides “Nothing”, 22.85% of images did not have the hand detected. While this is problematic, the live classification was simply adjusted so that a hand not being detected in the frame did not impact what it claimed the previous character to be, so it was an issue that was easily worked around. This model was tested on the dataset from the other project’s GitHub, and due to this issue it was never able to detect hands due to how the images were cropped, which removed key structure from the image that MediaPipe uses to detect hands. However, with the testing dataset that was initially used, all of these models far exceeded the baseline accuracy of 85% that was the initial goal. This model was able to generalize incredibly effectively, so it was selected for the live videofeed classification model.

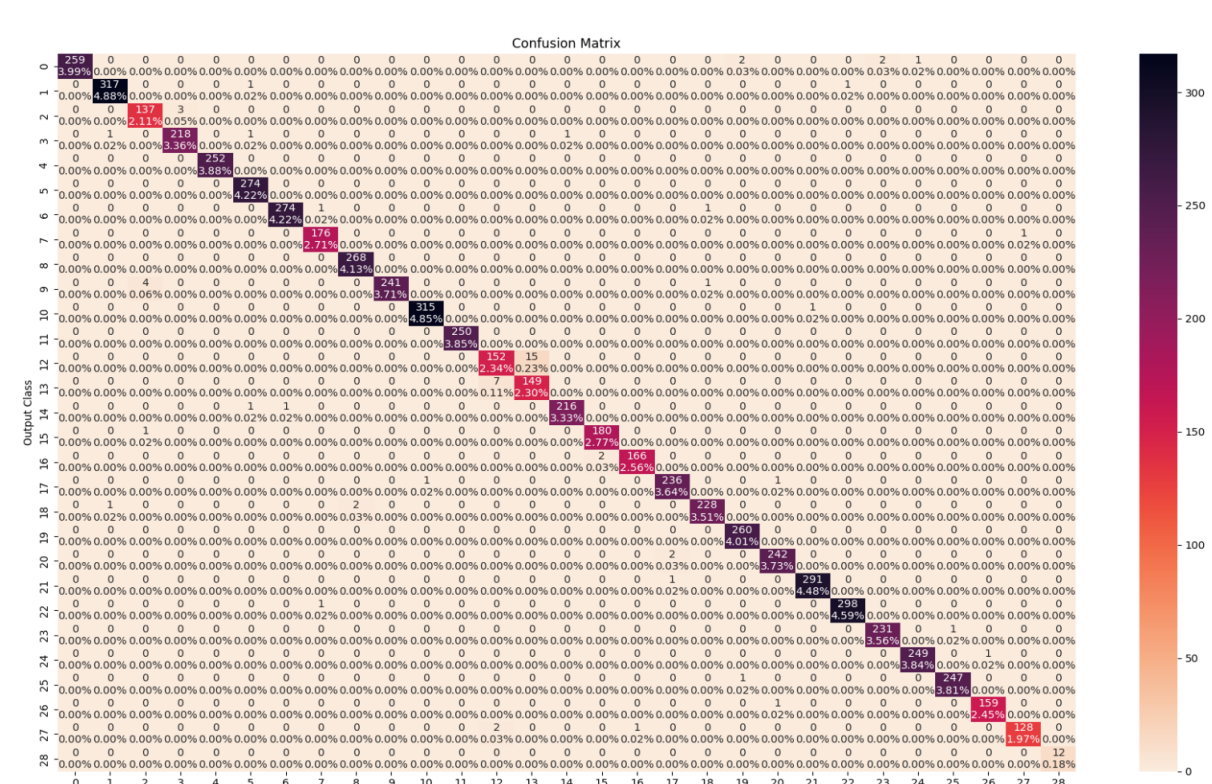

**Figure 9:** MediaPipe Model Confusion Matrix

In the testing of the live program, each letter was gone through and was attempted to be typed 10 times in rapid succession. There are some subtleties to the use of this program, as specific letters such as ‘M’ and ‘N’, ‘U’ and ‘V’, or ‘D’ and ‘O’ have very similar hand positioning–often differing by the slight rotation of a hand or the positioning of one finger.

Following this testing scheme, the confusion matrix seen in Figure X was produced. Clearly, someone with a bit of experience using the program can achieve outstanding results. Overall, there was a total accuracy of 92.14%. This is a very successful result for a generalized program with a provenly functional delete operation.

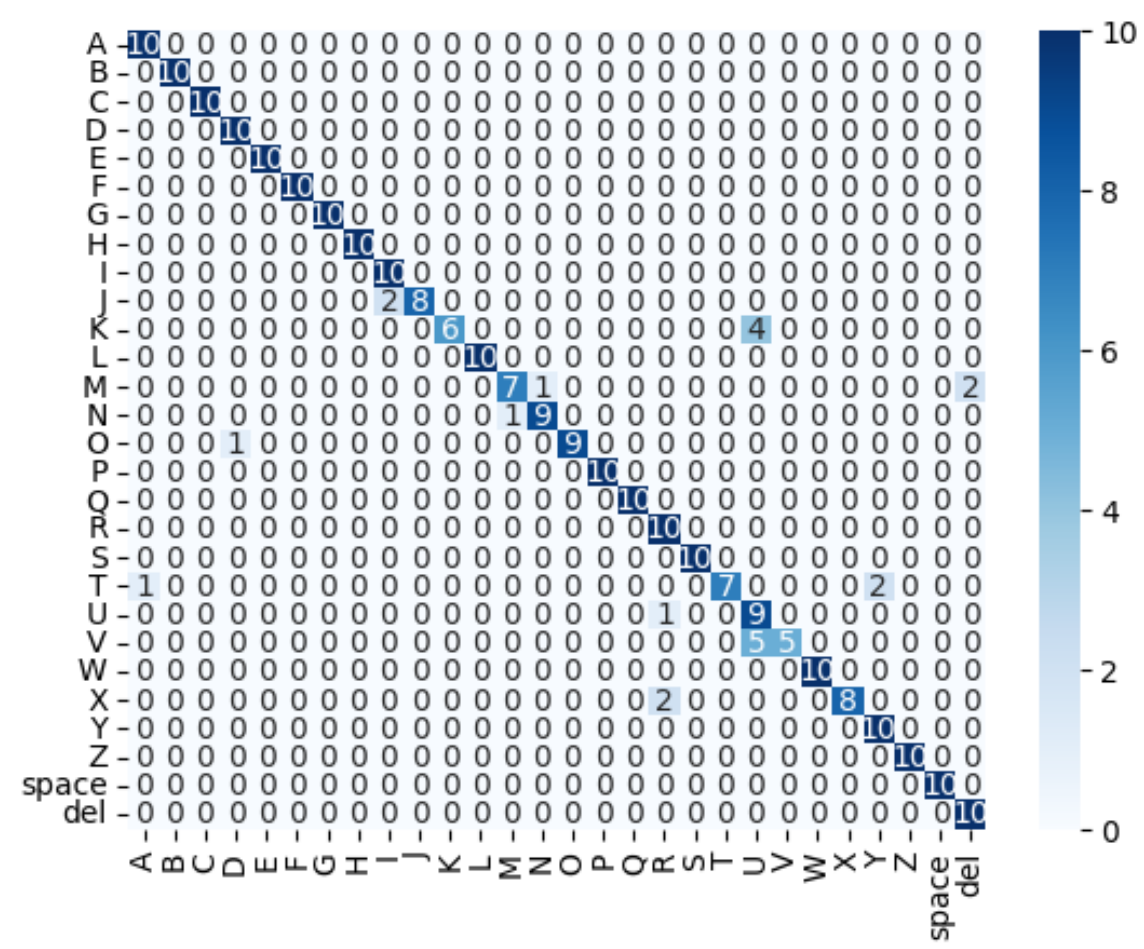

**Figure 10:** Live Testing Confusion Matrix

A particular point of confusion came for the letter ‘V’ being classified as ‘U’. This appeared to amount to the sufficient separation of the fingers to be detected as a ‘V’. However, even with this known, it was difficult to achieve remarkable results. Letters like ‘A’, ‘B’, ‘C’, and ‘L’ which have distinct unique hand positions were easily classified correctly.

## Discussion:

This project set out to create a neural network-based approach to the live classification of hand gestures in a live video feed with accuracy that is highly generalizable to a variety of settings and people. A great amount of success in doing so can be attributed to the use of the MediaPipe library for hand joint detection. This splitting of hand detection and hand pose classification into two tasks alleviated a lot of the issues that can come with having sufficient data to generalize something as complex as hand positioning.

Despite the significant success in completing this project, there are ways that the project can be improved. One big way that the project could be improved from a usability perspective is the expansion of the training set into more commonly used sign language words to better align with what ASL enables. This would be a large leap in overall usability, but it would be at the cost of needing a larger bank of data to train on.

Another improvement that can be made is in regards to the use of consecutive letters. The existing application requires that the hand be taken out of frame before being re-introduced as a new letter. This is no easy feat to overcome, which is why it was considered out of scope for this project. Either ingenious approaches in hard coded problems or predictive models that take the time between signs into account would be necessary to achieve a smooth running program of high accuracy.

Another improvement that could be made with more time is some more fine tuning. Certain letters are more likely to be misclassified than others, such as “U” and “V” mentioned above. A secondary neural network could be trained on just those letters, and if the first network returns either one, the secondary network would be called on to differentiate them since it would be more specialized. This would take a substantial amount of time to do for every group of letters that had misclassification issues, but it would likely improve accuracy.

A final way that this can be improved is via an accessibility point of view. The current program must be executed as a Python program. Setting out to put this into a web app so that users of any skill level can reap its benefits would be a key step toward making a program based on accessibility more accessible. The GitHub repository for this project can be found at https://github.com/BWurster/cs539-sp2023.